\documentclass[runningheads]{llncs}

\usepackage{eccv}

\usepackage{eccvabbrv}

\usepackage{graphicx}
\usepackage{booktabs}

\usepackage[accsupp]{axessibility}  %

\usepackage{graphicx}	
\usepackage{amsmath}	
\usepackage{amssymb}
\usepackage{booktabs}
\usepackage{times}
\usepackage{microtype}
\usepackage{caption}
\usepackage{float}
\usepackage{placeins}
\usepackage{color, colortbl}
\usepackage{stfloats}
\usepackage{enumitem}
\usepackage{tabularx}
\usepackage{xstring}
\usepackage{multirow}
\usepackage{xspace}
\usepackage{url}
\usepackage{subcaption}
\usepackage{xcolor}
\usepackage{pifont}
\usepackage[hang,flushmargin]{footmisc}
\graphicspath{{figs/}}

\usepackage{hyperref}

\usepackage{orcidlink}

\begin{document}

\newcommand{\dataset}{\textsc{DarkClusters-15k}}
\newcommand{\datasetsize}{15,000}
\newcommand{\los}{r}
\newcommand{\losdir}{\vec{r}}
\newcommand{\bbeta}{\boldsymbol{\beta}}
\newcommand{\btheta}{\boldsymbol{\theta}}
\newcommand{\balpha}{\boldsymbol{\alpha}}
\newcommand{\bgamma}{\boldsymbol{\gamma}}
\newcommand{\boldeta}{\boldsymbol{\eta}}
\newcommand{\bxi}{\boldsymbol{\xi}}
\newcommand{\band}{f}
\newcommand{\Ib}{\textbf{P}_\band}

\newcommand{\new}[1]{#1}
\newcommand{\nbf}[1]{{\noindent \textbf{#1.}}}
\newcommand{\supp}{Supplemental}

\newcommand{\todo}[1]{{\textcolor{red}{[TODO: #1]}}}

\newcommand{\R}[1]{{%
    \textbf{%
        \ifstrequal{#1}{1}{\textcolor{red}{R#1}}{%
        \ifstrequal{#1}{2}{\textcolor{blue}{R#1}}{%
        \ifstrequal{#1}{3}{\textcolor{magenta}{R#1}}{%
        \ifstrequal{#1}{4}{\textcolor{teal}{R#1}}{%
                           \textcolor{cyan}{R#1}%
        }}}}%
    }%
}}

\newcommand{\fref}[1]{Figure~\ref{#1}}
\newcommand{\tref}[1]{Table~\ref{#1}}
\newcommand{\eref}[1]{Equation~\ref{#1}}
\newcommand{\eeref}[2]{Equations~\ref{#1} and \ref{#2}}
\newcommand{\sref}[1]{Section~\ref{#1}}
\newcommand{\ssref}[2]{Sections~\ref{#1}~and~\ref{#2}}
\newcommand{\aref}[1]{Annex~\ref{#1}}

\definecolor{red}{rgb}{0.8,0,0}
\definecolor{purered}{rgb}{1,0,0}
\definecolor{pink}{rgb}{0.9,0,0.9}
\definecolor{darkred}{rgb}{0.6,0,0}
\definecolor{green}{rgb}{0.0,0.5,0}
\definecolor{blue}{rgb}{0,0,0.75}
\definecolor{darkblue}{rgb}{0,0,0.55}
\definecolor{lightcyan}{rgb}{0.5,0.7,0.7}
\definecolor{orange}{rgb}{0.9,0.3,0.1}
\definecolor{purple}{rgb}{0.6,0.0,0.6}
\definecolor{cyan}{rgb}{0.0,0.7,0.7}
\definecolor{darkgray}{rgb}{0.4,0.4,0.4}
\definecolor{bronze}{rgb}{0.7, 0.4, 0.18}
\definecolor{dorange}{rgb}{0.75, 0.4, 0.0}
\definecolor{darkgray}{rgb}{0.25,0.25,0.25}
\definecolor{black}{rgb}{0.0,0.0,0.0}

\newcommand{\comment}[3][magenta]{\textcolor{#1}{\emph{(\textbf{#2}: #3})}}
\newcommand{\diegoc}[1]{\comment[purple]{Diego}{#1}}
\newcommand{\diego}[1]{\comment[purple]{Diego}{#1}}
\newcommand{\D}[1]{\comment[purple]{Diego}{#1}}
\newcommand{\droyo}[1]{\comment[blue]{Droyo}{#1}}
\newcommand{\adolfo}[1]{\comment[orange]{Adolfo}{#1}}
\newcommand{\katie}[1]{\comment[red]{Katie}{#1}}
\newcommand{\brandon}[1]{\comment[bronze]{Brandon}{#1}}

\newcommand{\customcitet}[2]{#1~et~al.~\cite{#2}}

\title{Mapping Dark-Matter Clusters via Physics-Guided Diffusion Models} 

\titlerunning{Mapping Dark-Matter Clusters via Physics-Guided Diffusion Models}

\author{Diego Royo\inst{1}\orcidlink{0000-0001-6880-322X} \and
Brandon Zhao\inst{2} \and
Adolfo Muñoz\inst{1}\orcidlink{0000-0002-8160-7159} \and
Diego Gutierrez\inst{1}\orcidlink{0000-0002-7503-7022} \and\newline
Katherine L. Bouman\inst{2}\orcidlink{0000-0003-0077-4367}}

\authorrunning{D. Royo, B. Zhao, A. Muñoz, D. Gutierrez, and K. Bouman}

\institute{Universidad de Zaragoza--I3A, Spain \and
California Institute of Technology, USA}

\maketitle
\vspace{-.175in}

\begin{abstract}

Galaxy clusters are powerful probes of astrophysics and cosmology through gravitational lensing: their mass, dominated by 85\% dark matter, distorts background light. Yet, mass reconstruction lacks the scalability and large-scale benchmarks to process the hundreds of thousands of clusters expected from forthcoming wide-field surveys. %
We introduce a fully automated method to reconstruct cluster surface mass density from photometry and gravitational lensing observables. Central to our approach is \dataset, our new dataset of \datasetsize\ \new{mock cluster observations} with paired mass and photometry maps, the largest to date, spanning multiple redshifts and simulation frameworks.
We train a plug-and-play diffusion prior on \dataset\ that learns the statistical relationship between mass and light, and draw \new{approximate} posterior samples constrained by weak- and strong-lensing observables, yielding principled reconstructions with well-calibrated empirical uncertainties.
Our approach requires no expert tuning, runs in minutes rather than hours, achieves higher accuracy, 
and matches expert-tuned reconstructions of the MACS 1206 cluster.
We release our method and \dataset\ to support upcoming wide-field cosmological surveys.

\vspace{-.05in}
\end{abstract}

\vspace{-.325in}
\section{Introduction}
\label{sec:intro}
\vspace{-.05in}

Studying galaxy clusters provides meaningful insights into the fundamental nature and formation of our Universe; their abundance and mass profiles place vital constraints on astrophysical and cosmological models. Because dark matter constitutes roughly 85\% of their mass, accurately mapping their total mass distribution is essential to connect this invisible dark matter with observables such as galaxy luminosities, and to enable techniques like time-delay cosmography to measure the Universe's expansion rate.

One of the most powerful ways to probe the mass of galaxy clusters is through \emph{gravitational lensing}--the bending of light from background sources as it passes through the cluster’s curved space-time. 
We target massive clusters, exceeding $10^{13.5}$ times the mass of our Sun $M_\odot$. In these systems, the dense cores give rise to strong gravitational lensing that can produce multiple images of the same object, while the outer cluster regions produce weak lensing that only subtly distorts shapes.
The rarest, most massive clusters of $10^{14\text{--}15} M_\odot$ provide abundant lensing constraints, hence most mass mapping studies focus on these systems \cite{abdelsalam1998nonparametric, meneghetti2017frontier, napier2023hubble, diego2025euclid}. In contrast, clusters of $10^{13.5\text{--}14} M_\odot$ are ten to hundreds of times more common but yield far fewer lensing observables. Mass mapping analysis of these smaller systems is thus highly ill-posed, and often relies on simplified parametric models which depend heavily on expert tuning \cite{Ding_2025}, or on stacking many systems to recover only ensemble-average properties \cite{Ingoglia_2022}.

We present two main contributions. First, we release \dataset, our dataset of \datasetsize\ \new{mock cluster observations} derived from IllustrisTNG \cite{Pillepich_2017} and SIMBA \cite{dave2019simba} with paired surface mass density and photometry information. It is the largest of its kind, enabling rigorous benchmarking of mass mapping methods. Second, we present our \textit{fully automated} method to estimate the non-parametric mass distribution of galaxy clusters. %
Since lensing information alone is insufficient to fully recover the underlying mass distribution, our method integrates a diffusion-based generative prior 
with gravitational lensing physics operators that ensure consistency with observations. Unlike other approaches that impose a rigid light-traces-mass (LTM) assumption, our prior captures more flexible and expressive dependencies by modeling the joint distribution of mass and light. 
By factorizing the prior and data likelihood, our approach establishes a physics-based principled framework.
We formulate reconstruction as a reverse diffusion process based on posterior sampling, 
allowing us to quantify per-pixel uncertainties.

\newcommand{\cGREEN}{\cellcolor[HTML]{AAFFAA}}
\newcommand{\cYELLOW}{\cellcolor[HTML]{FFFFAA}}
\newcommand{\cRED}{\cellcolor[HTML]{FFAAAA}}
\newcommand{\cNO}{\cRED \ding{55}}
\newcommand{\cCHECKLTM}{\cYELLOW \checkmark (LTM 1:1)}
\newcommand{\cCHECK}{\cGREEN \checkmark}

\begin{table*}[t]
\centering
\setlength{\tabcolsep}{2.3pt} %
\renewcommand{\arraystretch}{1.15} %
\caption{Overview of galaxy cluster mass reconstruction methods. We differentiate between tools that require expert configuration (e.g., Lenstool \cite{kneib2011lenstool}, WSLAP+ \cite{diego2007combined}) and fully automatic approaches (MARS \cite{cha2022mars}, the method from work by \customcitet{Napier}{napier2023hubble}, and ours). Our free-form approach leverages a diffusion prior trained on \dataset, enabling 512×512 free parameters with mass reconstructions within minutes, significantly faster than traditional methods. Unlike other methods that assume that Light Traces Mass (LTM) with a 1:1 ratio, we freely model photometry along with Strong Lensing (SL) and Weak Lensing (WL). \emph{\textsuperscript{*}Our reimplementation.}
}
\vspace{-5pt}
\label{tab:comparison}
\begin{tabular}{|l|l|r|lll|l|l|}
\hline
\multicolumn{1}{|c|}{\multirow{2}{*}{\textbf{Method name}}} &
\multicolumn{1}{c|}{\multirow{2}{*}{\begin{tabular}[c]{@{}c@{}}\textbf{Model}\\[-3pt] \textbf{type}\end{tabular}}} &
\multicolumn{1}{c|}{\multirow{2}{*}{\textbf{\# Param.}}} &
\multicolumn{3}{c|}{\textbf{Inputs}} &
\multicolumn{1}{c|}{\multirow{2}{*}{\begin{tabular}[c]{@{}c@{}}\textbf{Automation}\\[-3pt] \textbf{level}\end{tabular}}} &
\multicolumn{1}{c|}{\multirow{2}{*}{\begin{tabular}[c]{@{}c@{}}\textbf{End-to-end}\\[-3pt] \textbf{time}\end{tabular}}} \\ \cline{4-6}
\multicolumn{1}{|c|}{} &
\multicolumn{1}{c|}{} &
\multicolumn{1}{c|}{} &
\multicolumn{1}{c|}{\textbf{Photometry}} &
\multicolumn{1}{c|}{\textbf{SL}} &
\multicolumn{1}{c|}{\textbf{WL}} &
\multicolumn{1}{c|}{} &
\multicolumn{1}{c|}{} \\ \hline
Lenstool \cite{kneib2011lenstool} & Parametric & $\sim$10--100 & \multicolumn{1}{l|}{\cCHECKLTM} & \multicolumn{1}{l|}{\cCHECK} & \cCHECK & \cRED Expert-guided & \cYELLOW Hours \cite{meneghetti2017frontier} \\ \hline
WSLAP+ \cite{diego2007combined} & Hybrid & $\sim$1000 & \multicolumn{1}{l|}{\cCHECK} & \multicolumn{1}{l|}{\cCHECK} & \cCHECK & \cRED Expert-guided & \cYELLOW Hours \\ \hline
MARS \cite{cha2022mars} & Free-form & 100$\times$100 & \multicolumn{1}{l|}{\cNO} & \multicolumn{1}{l|}{\cCHECK} & \cCHECK & \cGREEN Fully automatic & \cYELLOW Hours \cite{cha2022mars} \\ \hline
\textls[-30]{\customcitet{Napier}{napier2023hubble}}\textsuperscript{*} & Parametric & $\sim$10--100 & \multicolumn{1}{l|}{\cCHECKLTM} & \multicolumn{1}{l|}{\cNO} & \cNO & \cGREEN Fully automatic & \cGREEN Minutes \\ \hline
Ours & Free-form & 512$\times$512 & \multicolumn{1}{l|}{\cCHECK} & \multicolumn{1}{l|}{\cCHECK} & \cCHECK & \cGREEN Fully automatic & \cGREEN Minutes \\ \hline
\end{tabular}
\vspace{-.2in}
\end{table*}

\let\cGREEN\relax
\let\cYELLOW\relax
\let\cRED\relax
\let\cNO\relax
\let\cCHECKLTM\relax
\let\cCHECK\relax

Compared to other approaches (\tref{tab:comparison}), our method presents many advantages: it is non-parametric, can handle complex cluster structures where the mass-to-light ratio does not follow a strict 1:1 relationship, does not require cluster-specific expert tuning, and provides results in minutes instead of hours \cite{cha2024precision, napier2023hubble}.
Moreover, we obtain better quantitative and qualitative results on held-out \dataset\ clusters from different redshifts and simulation frameworks. We also reconstruct the MACS 1206 cluster, qualitatively matching expert-level reconstructions.

Our method is timely: upcoming sky surveys with improved telescope optics are expected to discover hundreds of thousands of clusters above $10^{13.5} M_\odot$ \cite{sartoris2016next, wen2024catalog, natarajan2024strong}\footnote{\href{https://www.esa.int/Science_Exploration/Space_Science/Euclid/Euclid_opens_data_treasure_trove_offers_glimpse_of_deep_fields}{For instance, the Euclid space telescope recently discovered 26 million galaxies in just one week.}}, with higher quality lensing data \cite{huff2025kinematic}. 
Automated, scalable methods like ours will be essential to analyze these systems without expert intervention, enabling statistically robust tests of cosmology models. By relaxing rigid mass-to-light assumptions, our approach could even automate the detection of unobservable dark matter. We release our \dataset\ dataset as a benchmark for future research, with code and further details in our project page\footnote{\url{https://graphics.unizar.es/projects/DarkMatterMapping/}}.

\section{Related work}
\label{sec:related}

\paragraph{Galaxy cluster mass reconstruction.}
Galaxy cluster mass is typically constrained using both \emph{strong} and \emph{weak} gravitational lensing. Strong lensing effectively probes the dense inner regions of the cluster, which can be represented reasonably well with parametric halo models \cite{kneib1996hubble, broadhurst2005strong, jullo2007bayesian} like NFW \cite{navarro1997universal} or PIEMD \cite{kassiola1993elliptic}.
Weak lensing provides wider spatial coverage but suffers from severe noise. The seminal work of Kaiser~and~Squires~\cite{kaiser1993mapping} introduced a direct inversion method, which has since been improved via sparsity-based regularization \cite{lanusse2016high, leonard2014glimpse}, inverse-variance filtering \cite{seljak1998weak}, or supervised deep learning \cite{jeffrey2020deep, remy2023probabilistic} to better handle the severe noise.
A complementary approach relies on the light-traces-mass assumption \cite{limousin2005constraining}, wherein parametric models scale mass according to observed luminosity \cite{kassiola1993elliptic, broadhurst2005strong}.

Modeling tools like Lenstool \cite{niemiec2020hybrid, kneib2011lenstool} and WSLAP+ \cite{diego2007combined} incorporate additional observables like photometry, X-ray or the Sunyaev-Zel'dovich effect. While these tools achieve state-of-the-art performance \cite{napier2023hubble, diego2025euclid},
both require expert guidance to define the number and type of parametric models for each cluster. We refer the reader to the comparison by \customcitet{Meneghetti}{meneghetti2017frontier} for more details. Alternatively, MARS \cite{cha2022mars,cha2024precision} is a free-form method combining strong lensing with a simple maximum-entropy regularizer.

By replacing rigid parametric assumptions and simple regularizations with a diffusion prior trained on \dataset, we capture an expressive mass-light joint distribution, constrained by weak and strong gravitational lensing, yielding non-parametric posterior estimates an order of magnitude faster than traditional optimizations.

\paragraph{Mass reconstruction using machine learning.}
With recent and upcoming surveys, both the quantity and quality of galaxy and galaxy-cluster observations are set to increase dramatically \cite{wen2024catalog, natarajan2024strong}. This rapid growth has positioned machine learning (ML) as a powerful tool for automated lens analysis.

Early ML work focused primarily on strong-lensing systems produced by individual galaxies typically only a few arcseconds across, employing convolutional neural networks trained on simulations to estimate model parameters \cite{hezaveh2017fast} and their uncertainties \cite{levasseur2017uncertainties}. %
Subsequent efforts have combined parametric lens models with more flexible, free-form components using image-space \cite{adam2022pixelatedreconstructiongravitationallenses}, wavelet-based \cite{galan2022using}, or neural implicit \cite{biggio2023modeling} representations, which has also been used to reconstruct unlensed background sources \cite{mishrasharma2022stronglensingsourcereconstruction}. Joint lens-and-source reconstruction has additionally been explored through variational inference \cite{rustig2024introducing} and diffusion models \cite{barco2025blind}.

Conversely, ML for weak lensing operates over much larger angular scales, leveraging constraints from vast galaxy populations spanning several degrees\footnote{1 degree is equivalent to 60 arcminutes and 3,600 arcseconds.} but at much coarser resolution.
In this regime, recent works utilize shear maps for diffusion-based posterior sampling \cite{boruah2025diffusion} or neural implicit representations for 3D mass inference \cite{zhao2024single, zhao2025revealing}.

Bridging these extremes, our method targets galaxy clusters, which typically span a few arcminutes, while maintaining sub-arcsecond resolution. Furthermore, our work is the first to combine a data-driven score prior for galaxy clusters with weak and strong lensing and photometry measurements at this scale. Thanks to our \dataset\ dataset, our model can learn fine structure in the joint mass/light distribution.%

\paragraph{Cosmological simulations and datasets.}
Cosmological simulations track the structural evolution of the Universe over cosmic time, capturing the interplay between baryonic (visible) and dark matter components at different redshifts. Each varies in volume, resolution, and cosmology assumptions. %
Projects like TNG-Cluster \cite{nelson2024introducing} or The Three Hundred project \cite{de2021three} isolate and simulate hundreds of rare, extremely massive clusters with virial masses $\ge10^{14}\,M_\odot$ at high resolution. Conversely, projects like MDPL2 \cite{klypin2016multidark}, Magneticum Pathfinder \cite{dolag2025encyclopedia} or BAHAMAS \cite{mccarthy2016bahamas} simulate larger regions, which include smaller clusters, but at coarser resolutions.

Our \dataset\ dataset is derived from the TNG300-1 dataset of the IllustrisTNG simulation suite \cite{Pillepich_2017}, and the SIMBA simulations \cite{dave2019simba}, which include many clusters in the $10^{13.5-14}\,M_\odot$ range, enabling us to extract \datasetsize\ pairs of surface mass density and photometry maps from galaxy clusters at sub-arcsecond resolution. %
Compared to other benchmarks such as the two hand-crafted clusters from the Frontier Fields project \cite{meneghetti2017frontier}, or the Camels Multifield Dataset \cite{villaescusa2022camels} which targets larger volumes at coarser resolution, \dataset\ enables large-scale benchmarking, being the largest dataset of its kind to date while also including our simulated gravitational lensing observations. Our work anticipates upcoming wide-field surveys that will discover many clusters in that mass range and their need for scalable reconstruction methods.

\section{Background}
\label{sec:background}

\subsection{Gravitational lensing}
\label{sec:background:lensing}

\begin{figure}[t]
    \centering
    \captionsetup{skip=4pt}
    \def\svgwidth{0.75\columnwidth} 
    \begin{small}
    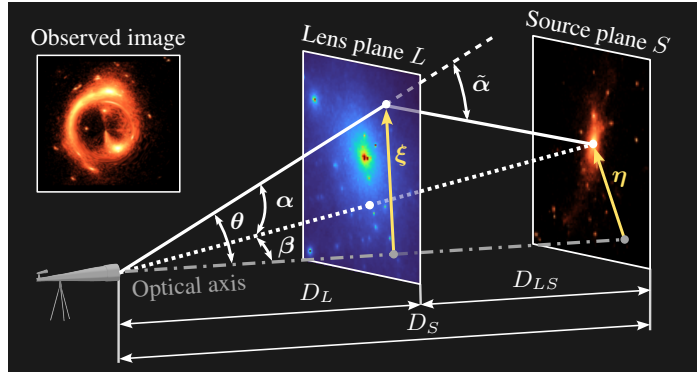
    \end{small}
    \caption{
    Sketch of a typical gravitational lens system. The light path of an object at angular position $\bbeta$ on a source plane $S$ is deflected by an angle $\tilde{\balpha}$ due to the presence of a galaxy cluster on the lens plane $L$, so the object appears at angular position $\btheta$ instead. 
    }
    \label{fig:grav_lensing_sketch}
    \vspace{-.2in}
\end{figure}

Here we outline the gravitational lensing formalism and refer the reader to \customcitet{Schneider} {schneider2006introduction} for details. \fref{fig:grav_lensing_sketch} illustrates an example, with a mass concentration at a distance $D_L$ and a light source at distance $D_S$ from the observer. \new{Given the vast distances involved, including the distance $D_{LS}$ between the mass and the source, we model them as \emph{lens} $L$ and \emph{source} $S$ \emph{planes} perpendicular to the optical axis.}
We will use $\btheta$ and $\bbeta$ for 2D \emph{angular positions} of objects in $L$ and $S$ with respect to the optical axis, and $\bxi \approx \btheta D_L$ and $\boldeta \approx \bbeta D_S$ for planar coordinates.

Light from the source at $\boldeta$ follows a curved path through the distorted space-time near the mass, thus we observe it at $\bxi$ instead. We simplify this continuous deflection into two straight rays bending by an angle $\tilde{\balpha}$ at the lens plane $L$, and establish from \fref{fig:grav_lensing_sketch}:
\begin{equation}
    \boldeta = \frac{D_S}{D_L}\bxi -  D_{LS}\,\tilde{\balpha}(\btheta)\; \implies \; \bbeta = \btheta - \balpha(\btheta),
    \label{eq:lens-equation}
\end{equation}
which simplifies to the expression on the right using angular positions and the \emph{reduced} deflection angle $\balpha(\btheta) \equiv \tilde{\balpha}(\btheta) D_{LS}/D_S$, and which is used in all our other derivations. %

The deflection $\balpha(\btheta)$ depends on the lens's surface mass density, denoted as $\Sigma(\btheta)$ and typically expressed via the dimensionless \emph{convergence} $\kappa(\btheta)$:
\begin{equation}
    \kappa(\btheta) = \frac{\Sigma(\btheta)}{\Sigma_{cr}}, \quad \Sigma_{cr}(D_L, D_S) = \frac{c^2}{4\pi G}\frac{D_S}{D_LD_{LS}},
    \label{eq:kappa}
\end{equation}
where $c$ and $G$ are the speed of light and the gravitational constant, respectively,
and $\Sigma_{cr}(D_L, D_S)$ is the critical surface mass density. %
Regions where $\Sigma \geq \Sigma_{cr}$ (i.e., $\kappa \geq 1$) \new{can} produce multiple images, defining the \emph{strong} lensing regime. Conversely, the \emph{weak} lensing regime measures small background distortions in outer regions where $\kappa \ll 1$.

\paragraph{Strong lensing regime.} We use \eref{eq:lens-equation}, obtaining the reduced deflection angle $\balpha(\btheta)$ from the convergence $\kappa(\btheta)$ as a convolution over all angular positions $\btheta'$ in the lens plane, computed efficiently in the Fourier domain:
\begin{equation}
    \balpha(\btheta) = \frac{1}{\pi} \int_{\mathbb{R}^2} \mathrm{d}^2\btheta' \, \kappa(\btheta') \, \frac{\btheta - \btheta'}{\left\vert \btheta - \btheta' \right\vert^2}.
    \label{eq:alpha}
\end{equation}

\paragraph{Weak lensing regime.} Here, deflections caused by the lens plane are small enough to linearize  $\balpha(\btheta)$ in \eref{eq:lens-equation} via its Jacobian matrix $\mathcal{A}$:
\begin{equation}
    \bbeta
    \approx
    \mathcal{A}(\btheta) \btheta,
    \quad
    \mathcal{A}(\btheta)
    = \begin{pmatrix} 1 - \kappa - \gamma_1 & -\gamma_2 \\ -\gamma_2 & 1 - \kappa + \gamma_1 \end{pmatrix}, 
    \label{eq:lens-equation-linear}
\end{equation}
where $\bgamma(\btheta) = \gamma_1(\btheta) + i\gamma_2(\btheta)$ is the complex-valued lens shear. Weak lensing assumes that all sources (i.e., background galaxies) have an elliptical morphology, thus \eref{eq:lens-equation-linear} maps their \emph{intrinsic} ellipticity (in the source plane) to their \emph{observed} ellipticity (in the lens plane) by altering their axis ratio and orientation.
Because intrinsic shapes are unknown, deriving shear from observed ellipticities is highly noisy, with a signal-to-noise ratio (SNR) of $\sim$0.01-0.1.
In our work, we compute shear from convergence as:
\begin{equation}
    \bgamma(\btheta) = \frac{1}{\pi} \int_{\mathbb{R}^2} \mathrm{d}^2\btheta' \, \kappa(\btheta') \, \mathcal{D}(\btheta - \btheta'), \qquad \mathcal{D}(\btheta) = \frac{\theta^2_2 - \theta^2_1 - 2i\theta_1 \theta_2}{\left\vert \btheta \right\vert^4}.
    \label{eq:shear}
\end{equation}
also using a convolution as in \eref{eq:alpha}, only now the kernel $\mathcal{D}$ uses the horizontal $\theta_1$ and vertical $\theta_2$ components of the angular position $\btheta$.

\subsection{Diffusion models}
\label{sec:background:diffusion}

Diffusion models provide a flexible framework for generative modeling by learning a score-based prior that captures the structure of complex data distributions.
This generative process works by reversing the forward diffusion process which iteratively perturbs a data sample by adding noise \cite{ho2020denoising}. Mathematically, the forward process $\{\textbf{x}_t\}_{t=0}^T$ starts with an image $\textbf{x}_0 \sim p_\text{data}$ from the data distribution $p_\text{data}$, and adds noise according to the following stochastic differential equation (SDE) \cite{song2021scorebasedgenerativemodelingstochastic}:
\begin{equation}
    \mathrm{d}\textbf{x}_t = f(\textbf{x}_t, t) \mathrm{d}t + g(t) \mathrm{d}\textbf{w},
    \label{eq:diffusion-forward}
\end{equation}
where $\textbf{w}$ represents n-dimensional Brownian motion. The drift and diffusion coefficients $f(\textbf{x}_t, t)$ and $g(t)$ parameterize the noise schedule. %

The reverse of \eref{eq:diffusion-forward}, which conversely transforms noisy images back to meaningful images in $p_\text{data}$, is another diffusion process given by the reverse-time SDE:
\begin{equation}
    \mathrm{d}\textbf{x}_t = \left[f(\textbf{x}_t, t) - g(t)^2 \nabla_{\textbf{x}_t}\log p(\textbf{x}_t)\right]\mathrm{d}t + g(t) \mathrm{d}\tilde{\textbf{w}},
    \label{eq:diffusion-reverse}
\end{equation}
where $\tilde{\textbf{w}}$ represents n-dimensional Brownian motion with reversed time, and $\mathrm{d}t$ is an infinitesimal negative timestep. Diffusion models learn the score function $\nabla_{\textbf{x}_t}\log p(\textbf{x}_t)$, after which \eref{eq:diffusion-reverse} generates new samples 
in $p_\text{data}$ starting from Gaussian noise. \new{We later build on posterior sampling methods~\cite{chung2022diffusion, zhang2025improving}
that augment this prior with a likelihood term that steers the reverse process towards solutions consistent with the data.}

\section{\dataset~dataset}
\label{sec:dataset}

The recent availability of large, high-fidelity cosmological simulations \cite{nelson2024introducing} has created a timely opportunity to build realistic datasets for cluster-scale lensing studies. Motivated by this development, we construct a new dataset that processes these large simulations to systematically isolate cluster environments and generate corresponding realistic lensing measurements. We additionally release the code used to create these simulations, enabling reproducibility and further expansion by the community.

\subsection{Dataset generation}
\label{sec:dataset:generation}

Our \dataset\ dataset contains \datasetsize\ \new{projected} galaxy cluster samples (split in 12,000 for training, 3,000 for test), derived from the IllustrisTNG \cite{Pillepich_2017} and SIMBA \cite{dave2019simba} cosmological simulations. It is the largest dataset of its kind. Each sample includes four $512 \times 512$ images: one surface mass density map $\Sigma(\boldsymbol\theta)$ and three simulated Hubble Space Telescope (HST) images $\Ib(\boldsymbol\theta)$ corresponding to wavelength filters $\band \in \{125, 606, 814\}$, representing near-infrared (F125W), and optical (F606W, F814W) bands. All images are centered on the cluster and are provided in two field-of-view settings: $100 \times 100$ and $225 \times 225$ arcseconds. The total gravitationally-bound mass for each cluster in \dataset\ is above $10^{13.5} M_\odot$, with most of them being below $10^{14} M_\odot$. See the \supp\ for the cluster distribution, units, and processing.

\begin{figure}[t]
  \centering
  \captionsetup{skip=-6pt}
  \def\svgwidth{\columnwidth} 
  \begin{small}
  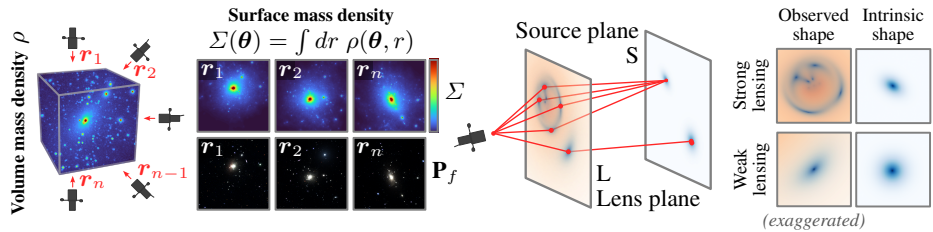
  \end{small}
  \caption{
  \textbf{Left:} For the creation of our \dataset\ dataset, we extract 600 simulated galaxy clusters from IllustrisTNG and SIMBA as 3D point clouds, and compute the surface mass density $\Sigma$ for $n = 25$ different line of sight $\vec{\los}_i$ projections. \textbf{Right:} For each projection, we then independently generate measurements for strong lensing (i.e., positions of multiple images for each source) and weak lensing (i.e., shear $\boldsymbol\gamma$ at the source positions).
  }
  \label{fig:dataset_creation}
  \label{fig:dataset_lensing}
\end{figure}

\paragraph{Data extraction and train/test splits.} \new{We extract 600 distinct 3D cluster simulations as point clouds, 540 from the TNG300-1 subset of IllustrisTNG and 60 from SIMBA. To augment the data, we project each cluster along 25 different viewing directions $\losdir_i$ (\fref{fig:dataset_creation}, left), which serve as lens planes $L$. Our training/validation set is formed exclusively from 480 TNG clusters (i.e., 12,000 projections) fixed at a distance $D_L$ (\fref{fig:grav_lensing_sketch}) corresponding to a redshift of $z_L = 0.5$. Our test set uses 120 clusters with two variations. First, we include 60 TNG clusters which are the same 20 clusters at three redshifts $z_L \in \{0.2, 0.5, 1.0\}$. This yields three TNG test sets of 500 projections each. Clusters at $z_L = 0.2$ are older, more massive clusters with complex merged structures, while clusters at $z_L = 1.0$ are younger and less evolved. Second, we use 60 clusters (1,500 projections) from SIMBA. Since SIMBA relies on different algorithms than TNG, this test set verifies our model's performance across different simulation frameworks.} %

\paragraph{Generating strong and weak lensing observations.} 
For every test dataset sample, we simulate both strong- and weak-lensing observations (\fref{fig:dataset_lensing}, right). These observations are used only at inference time and are not part of the diffusion-model training set. For strong lensing, we uniformly sample source angular positions $\bbeta$ on the source plane $S$ and solve \eref{eq:lens-equation} with \eref{eq:alpha} to identify configurations in which a single background galaxy is lensed into multiple, distinct image angular positions $\btheta$ in the lens plane $L$.
Depending on the cluster total mass, we simulate between $5$ and $20$ sources whose background galaxies are lensed into multiple distinct images, typical for clusters of $10^{13.5} M_\odot$ and above. For weak lensing, we compute the shear $\bgamma(\btheta)$ at each angular position $\btheta$ using \eref{eq:shear}. We adopt realistic source densities of 30 weakly lensed galaxies per arcmin$^2$ \cite{lanusse2016high} and add Gaussian noise following prior work \cite{huff2025kinematic, lanusse2016high} with variance $\sigma_\text{w}^2 \in \{0.03, 0.3\}$. The larger value corresponds to traditional weak lensing observations, where noise comes from the uncertainty in the galaxy's intrinsic shape, while the smaller value corresponds to kinematic observations, which reduces this shape noise by including velocity information. Since $\bgamma$ is small, these correspond roughly to SNRs of $0.01$ and $0.1$, respectively. For both regimes, we distribute sources throughout redshifts $z_S$ following a probability distribution $p(z_S) \propto \exp(-z_S/z_0) z_S^2/(2 z_0^3) $ \cite{lanusse2016high}. We use $z_0=2/3$ which yields a median redshift of $1.75$, and truncate at $z_S = 5$.

\section{Method}
\label{sec:method}

Our method reconstructs surface mass density maps of individual galaxy clusters using a physics-guided diffusion framework. We train a diffusion prior on \dataset, which provides realistic cluster mass and photometry distributions and enables the model to capture both their statistical structure and their mutual correlations (\sref{sec:method:prior}). To incorporate physical observations, we pair this prior with forward models for strong and weak gravitational lensing (\sref{sec:method:lensing}), yielding a hybrid approach that automatically produces high-fidelity reconstructions within minutes while simultaneously quantifying uncertainty.

To perform inference, we adapt the Decoupled Annealing Posterior Sampling (DAPS) 
algorithm~\cite{zhang2025improving}, as implemented in the InverseBench 
framework~\cite{zheng2025inversebench}. Our goal is to estimate the surface mass 
density $\Sigma$ (\sref{sec:background:lensing}), represented by the diffusion 
variable $\mathbf{x}_t$ at timestep $t$ (\sref{sec:background:diffusion}). 
DAPS enables efficient posterior sampling by combining a data-driven prior with the likelihood given by physics-based lensing models, ultimately yielding mass maps that are both physically plausible and tightly constrained by the observed lensing data.

To generate these samples, DAPS alternates between two different updates. Consider the marginal update from time $t + \Delta t$ to time $t$, which will update the value of $\textbf{x}_t$. In the first step, DAPS samples $\tilde{\mathbf{x}}_{0|\mathbf{y}}$:
\begin{equation}
\tilde{\mathbf{x}}_{0|\mathbf{y}} \sim p(\mathbf{x}_0 \mid \mathbf{x}_{t + \Delta t}, \mathbf{y}) \propto p(\mathbf{y} \mid \mathbf{x}_0)p(\mathbf{x}_0 \mid \mathbf{x}_{t + \Delta t}),
\label{eq:bayes}
\end{equation}
from a conditional posterior that combines the likelihood term $p(\mathbf{y} \mid \mathbf{x}_0)$, which enforces consistency with the observed lensing measurements, with the prior term $p(\mathbf{x}_0 \mid \mathbf{x}_{t+\Delta t})$ derived from the diffusion model, which encourages $\mathbf{x}_0$ to follow the learned distribution of realistic clusters. In the second step, it samples:
\begin{align}
\mathbf{x}_t \sim p(\mathbf{x}_{t} \mid \tilde{\mathbf{x}}_{0|\mathbf{y}} ) = \mathcal{N}( \tilde{\mathbf{x}}_{0|\mathbf{y}} 
,\, \sigma_{t}^2 \mathbf{I}),
\end{align}
injecting the Gaussian noise with variance $\sigma^2_t$. %
This two-step procedure is repeated for progressively smaller values of $t$ until $\mathbf{x}_0$ \new{approximates a sample from the target posterior distribution.}%

We now describe how observational constraints enter this posterior sampling. 
The weak- and strong-lensing observations are linked to the underlying mass 
distribution through well-defined physical forward models, so they naturally 
appear in the likelihood term $p(\mathbf{y} \mid \mathbf{x}_0)$. In contrast, 
the photometric data $\Ib$ does not admit a simple analytic relationship 
to the mass field. To incorporate this information, we augment the 
likelihood-driven constraints with a diffusion-model prior that is conditioned 
on photometry, allowing the model to learn the statistical relationship between 
light and mass directly from training data rather than imposing an explicit 
parametric form.

\subsection{Photometry-conditioned diffusion prior}
\label{sec:method:prior}

We train a DDPM++ network \cite{song2021scorebasedgenerativemodelingstochastic} to learn the diffusion prior used in DAPS, modeling realistic surface–mass-density distributions from \dataset. The network learns to predict the (unconditional) score $\nabla_{\textbf{x}_t}\log p(\textbf{x}_t)$ in \eref{eq:diffusion-reverse}, at timestep $t$ in the generation. Importantly, although DAPS traditionally uses an unconditional diffusion prior, we condition the score network on the corresponding photometric images $\Ib$, which are typically highly correlated with the mass maps.
This conditioning enables the model to learn $\nabla_{\mathbf{x}_t}\log p(\mathbf{x}_t \mid \Ib)$, injecting photometric information that cannot appear in the likelihood and guiding generation toward mass distributions consistent with the light.
We train the DDPM++ network using the EDM loss from Karras et al.~\cite{karras2022elucidating}. 
The network receives a four-channel image of mass and photometry where all channels are normalized to the range $[0, 1]$. For additional details about training parameters, see the \supp.

\subsection{Lensing likelihood}
\label{sec:method:lensing}

In order to introduce weak and strong gravitational lensing observations into our generation pipeline, we include multiple physical \emph{forward models} that we use to compute the likelihood $p(\mathbf{y}\mid \mathbf{x}_0)$ in \eref{eq:bayes} for observations $\textbf{y}$ given our mass density estimate $\textbf{x}_0$ obtained using our diffusion prior.
Specifically, we use the gradient of the log-likelihood $\nabla_\textbf{x} \log p(\mathbf{y}\mid \mathbf{x}_0)$ to update our estimate. Rather than modeling the exact likelihood, we formulate this term as proportional to a sum of $k$ independent loss terms $\mathcal{L}_k$  derived from our physical forward operators, providing a practical way to integrate gravitational lensing data into our pipeline. Each term is weighted by $\lambda_k$:
\begin{equation}
  -\log p(\mathbf{y}\mid \mathbf{x}_0) \propto \sum_k \lambda_k \mathcal{L}_k,
  \label{eq:losses}
\end{equation}
In the rest of this section, we will use a hat, such as in $\hat{\textbf{y}}_k$, to refer to our estimations, and no hat, such as in $\textbf{y}_k$, to refer to the observed data.

\paragraph{Strong lensing forward model.} In strong lensing, light from a background galaxy source on $S$ at angular position $\bbeta$ is deflected at multiple angular positions $\btheta$ on the lens plane $L$.
The true angular position $\bbeta$ is unknown, but the lens equation (\eref{eq:lens-equation}) must hold.
We compute our estimation $\hat{\bbeta}_i$ of the angular position of the $i$-th source on $S$:
\begin{equation}
  \hat{\bbeta}_i = \frac{1}{m_i}\sum_{j=1}^{m_i} \hat{\bbeta}_{i,j}, \;\; \text{where} \;\; \hat{\bbeta}_{i,j} = \btheta_{i,j} - \hat{\balpha}(\btheta_{i,j}), \quad 
\end{equation}
as the average of individual estimations $\bbeta_{i,j}$ from all $m_i$ images of the source.
We compute $\hat{\balpha}$ through \eeref{eq:kappa}{eq:alpha} using the current estimation of the surface mass density $\hat{\Sigma}$ and the known source distance $D_S$. We formulate our \emph{geometric} loss as the average squared error of all $n$ sources:
\begin{equation}
  \mathcal{L}^\text{geo}_\text{s} = \frac{1}{n} \sum_{i=1}^n \frac{1}{m_i} \sum_{j=1}^{m_i} \left\vert \hat{\bbeta}_{i,j} - \hat{\bbeta}_i \right\vert^2. 
\end{equation}
To prevent all images of all sources from collapsing to one unique point, we found it effective to incorporate photometry, comparing the observed image $\Ib(\btheta)$ to our estimation $\hat{\textbf{P}}_\band(\btheta; \hat{\bbeta})$ of the light at $\btheta$ from a light source placed at $\hat{\bbeta}$. Concretely, we model the light emission by this source using a parametric ellipsoidal (Sérsic) source \cite{sersic1963influence} at angular position $\hat{\bbeta}_i$. This allows us to introduce a \emph{photometric} loss:
\begin{equation}
  \mathcal{L}^\text{img}_\text{s} =\! \frac{1}{n} \sum_{i=1}^n \frac{1}{m_i}\! \sum_{j=1}^{m_i} \!\left( \Ib(\btheta_{i, j}) -\hat{\textbf{P}}_\band(\btheta_{i, j}; \hat{\bbeta_i}) \right)^2\!\!.
\end{equation}

\paragraph{Weak lensing forward model.} We can compute our estimated shear $\hat{\bgamma}$ through \eeref{eq:kappa}{eq:shear}. \new{Since weak lensing happens in the $\kappa \ll 1$ regime, we approximate reduced shear $g = \bgamma / (1 - \kappa) \approx \bgamma$ in our computations.} However, naive comparison between $\hat{\bgamma}$ and $\bgamma$ for all angular positions $\btheta$ is not possible for two reasons:
(i) these measurements come from sources at different distances $D_S$, thus from \eref{eq:kappa} we would need to compute $\kappa$ and later $\bgamma$ for each source, and
(ii) measurement is only possible at sparse angular positions $\btheta$ with a source at $S$.

To solve these issues, we can compensate for the difference in distances $D_S$ by scaling all observations to an arbitrary reference distance $D_R$ by multiplying and dividing by the corresponding critical densities $\Sigma_\text{cr}$.
Finally, we extrapolate sparse shear observations to the full field of view using a radial basis function (RBF) interpolator $\Phi(\bgamma, \btheta)$, yielding a dense shear map %
that covers the \new{lens plane $L$} from sparse values $\bgamma$. \new{Rather than an exact per-pixel noise model, this yields a smoothed data-fidelity term.} This step mimics standard weak-lensing smoothing methods \cite{lanusse2016high, kaiser1993mapping}, which suppress high-frequency noise while preserving the useful, large-scale part of the signal. It also enables computing data-fidelity gradients across the entire image rather than only at the observed points.
We combine all these steps in our weak lensing loss:
\begin{equation}
  \mathcal{L}_\text{w} = \frac{1}{ \sigma_\text{w}^2} \! \sum_{\btheta} \left\vert\hat{\bgamma}(\btheta) - \Phi\left(\bgamma \frac{\Sigma_\text{cr}(D_L, D_S)}{\Sigma_\text{cr}(D_L, D_R)}, \btheta \right) \right\vert^2.
\end{equation}
Shear measurements are very noisy; as per \sref{sec:dataset:generation}, we use values $\sigma_\text{w}^2 = 0.3$ and $\sigma_\text{w}^2 = 0.03$ depending on the observation regime.

\vspace{-.1in}
\section{Results and evaluation}
\label{sec:results}

Here we analyze the performance of our physically guided diffusion method for galaxy-cluster mass estimation. In \sref{sec:results:uncertainty} we show results of our posterior sampling and verify that our model's predicted uncertainty is well calibrated. In \sref{sec:results:ablation}, we leverage our principled framework to conduct an ablation study, evaluating our method's performance using different combinations of inputs. Then, to provide fair and meaningful baselines, in \sref{sec:results:ourdataset} we compare against three automated approaches: \new{Kaiser-Squires inversion \cite{kaiser1993mapping}, which relies solely on weak-lensing constraints;} our implementation of the approach from \customcitet{Napier}{napier2023hubble}, which operates directly on photometric images $\Ib$ with comparable runtime; and our UNet model baseline, which maps photometry $\Ib$ directly to surface mass density. In contrast, many of the methods listed in \tref{tab:comparison} either require substantial, cluster-specific expert tuning (e.g., Lenstool or WSLAP+) or lack publicly available implementations (e.g., MARS \cite{cha2022mars}). Finally, in \sref{sec:real-cluster} we apply our method to reconstruct the surface mass distribution of the real MACS 1206 cluster and compare it to an existing manual, expert-tuned reconstruction.

\subsection{Posterior sampling and uncertainty calibration}
\label{sec:results:uncertainty}

\begin{figure}[t]
    \centering
    \captionsetup{skip=-6pt}
    \def\svgwidth{\columnwidth} 
    \begin{small}
    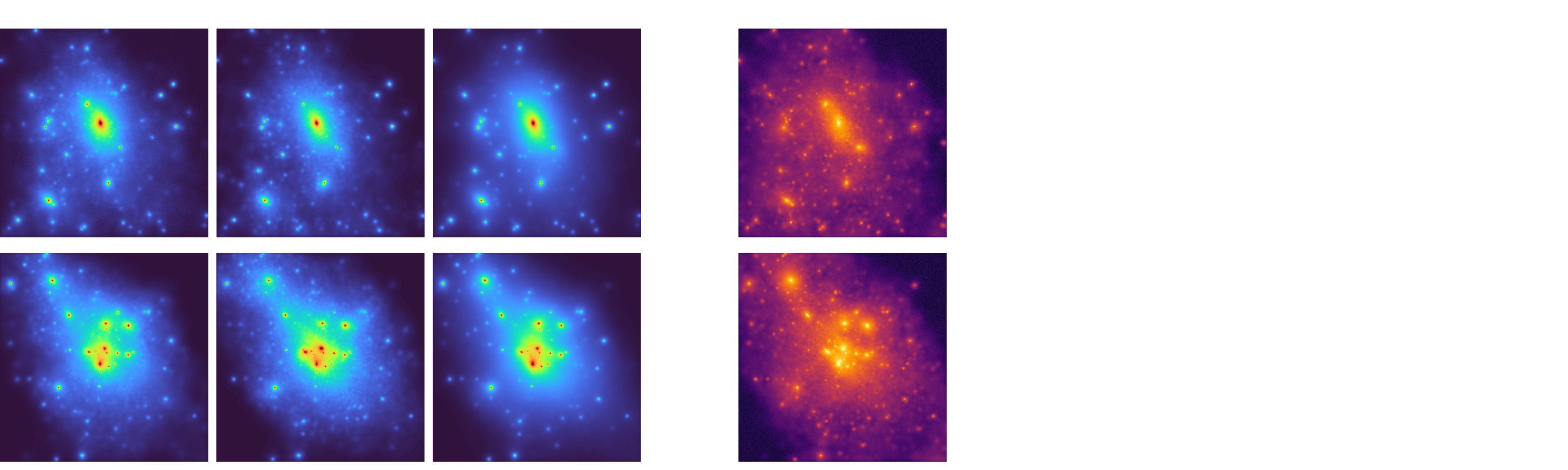
    \end{small}
    \caption{
    \textbf{Left:} Our diffusion-based approach produces multiple plausible reconstructions; in each row we show one representative sample of the lens convergence $\kappa$, and the posterior mean and standard deviation over 20 samples. \textbf{Right:} Reliability diagram, showing close correlation between our model's predicted uncertainty and the actual error. %
    }
    \label{fig:uncertainty}
\end{figure}

Our model can sample a range of plausible mass density maps. In \fref{fig:uncertainty}, left, each row has one example for which we show one representative posterior sample, and the mean and standard deviation over 20 samples. The variance across these samples captures the range of plausible mass configurations consistent with the measurements and diffusion prior, with higher variance indicating regions where the data weakly constrain the solution. We verify that our predicted uncertainty closely tracks the actual reconstruction error. \fref{fig:uncertainty}, right, \new{shows a reliability diagram using the procedure from work by \customcitet{Kuleshov}{kuleshov2018accurate}, which compares observed against expected confidence levels on a held-out test set. The agreement between the two indicates our marginal coverage is largely accurate, and suggests that our model's uncertainty is well calibrated.}

\subsection{Ablation: effect of photometry and strong/weak lensing}
\label{sec:results:ablation}

\begin{figure}[t]
    \centering
    \captionsetup{skip=-6pt}
    \def\svgwidth{\columnwidth} 
    \begin{small}
    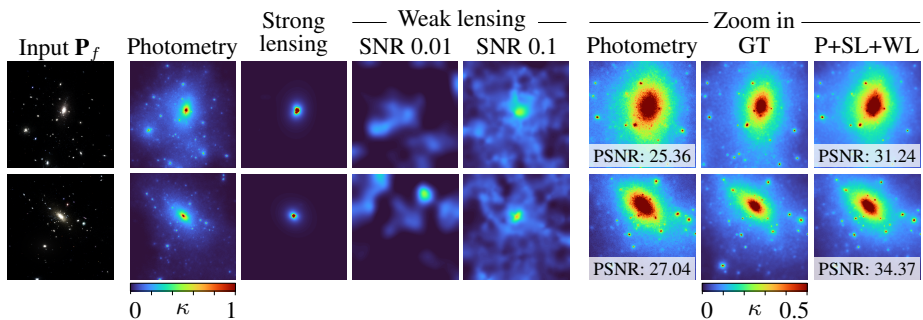
    \end{small}
    \caption{Ablation study on the different input components used by our method to estimate cluster masses. \textbf{Left:} Each row shows photometry $\Ib$ for one cluster, and mass reconstructions using only one input as stated. \textbf{Right:} Adding strong (SL) and weak lensing (WL) information (SNR 0.01) improves the photometry (P) results. We report PSNR of each image w.r.t. the ground truth (GT).}
    \label{fig:ablation-tests}
    \vspace{-.1in}
\end{figure}

In \fref{fig:ablation-tests}, we examine how different inputs to our pipeline contribute to the final reconstruction. Referring to \eref{eq:bayes}, we 
disentangle the influence of the learned prior $p(\mathbf{x}_t)$---which 
captures the joint distribution of mass and photometry---from that of the 
likelihood terms $p(\mathbf{y}\mid \mathbf{x}_0)$, which arise from the physical 
forward models for strong and weak lensing. In the leftmost column, we show an RGB representation of the photometry $\Ib$ of the cluster as reference, while the remaining columns show reconstructions using a different single input. 
While photometric information $\Ib$ alone provides a useful initial estimate, it cannot completely account for all dark matter as it does not interact with light. Thus, strong and weak gravitational lensing observables are necessary to fully reconstruct the cluster's surface mass density distribution. Strong lensing, in particular, constrains the surface mass density 
near the cluster core. Weak lensing contributes more subtly due to noise (with an SNR of $0.01\text{-}0.1$) and sparsity ($\sim$420 observations is typical for a $225\times225$ field of view), but still provides valuable information about the outer regions of the cluster. As shown in \fref{fig:ablation-tests}, right, relying solely on photometry can cause the estimation to overshoot. Adding strong and weak lensing constraints successfully corrects this issue, also improving the computed PSNR by more than five points.

\subsection{Baseline comparisons}
\label{sec:results:ourdataset}

\new{We compare against three baselines. First, we compare against Kaiser-Squires inversion \cite{kaiser1993mapping}, which relies on weak-lensing signal alone. This approach performs poorly as these constraints are very sparse and noisy. For reference, we also show our strong/weak-lensing-only reconstructions in \sref{sec:results:ablation}. Second,} we implement the approach from \customcitet{Napier}{napier2023hubble}, which employs the publicly available SE++ tool \cite{bertin1996sextractor} to estimate the surface mass density of observed galaxy clusters. Their work proceeds in two steps. First, they use SE++ to extract the positions, shapes and orientations of elliptical light sources from a photometric image $\Ib$ (we use the wavelength filter $\band = 125$ for this). These light sources 
correspond to the galaxy members of the cluster. Second, they estimate the cluster's mass distribution by summing multiple parametric mass components, each modeled with a PIEMD profile \cite{kassiola1993elliptic} and assigned to each of the extracted galaxy members.
The mass assigned to each parametric model has a 1:1 ratio with respect to light, following the relations discussed in \customcitet{Limousin}{limousin2005constraining}. Finally, we also include our UNet baseline, which directly maps photometry $\Ib$ to surface mass density. 
Note that neither Napier et al. nor UNet account for gravitational lensing observables, so they can only infer the invisible dark matter indirectly from luminous sources. In contrast, our approach integrates gravitational lensing observables to reconstruct the complete mass distribution.

\begin{table}[t!]
\centering
\setlength{\tabcolsep}{2.5pt} %
\renewcommand{\arraystretch}{1.25} %
\caption{Evaluation results for galaxy clusters at $z_L = 0.5$ in the TNG and SIMBA test sets of \dataset. We report the Peak Signal-to-Noise Ratio (PSNR), Structural Similarity Index (SSIM), Pearson Correlation Coefficient (PCC), and the $\kappa$, $\balpha$ and $\bgamma$ RMSE at held-out points (i.e., points without any photometry or lensing signals).}
\vspace{-5pt}
\label{tab:quantitative-results}
\begin{tabular}{|l|rr|rr|rr|rr|}
\hline
\multirow{2}{*}{\textbf{Metric}} & \multicolumn{2}{c|}{\textbf{KS}~\cite{kaiser1993mapping}} & \multicolumn{2}{c|}{\textbf{Napier et al.}~\cite{napier2023hubble}} & \multicolumn{2}{c|}{\textbf{UNet}} & \multicolumn{2}{c|}{\textbf{Ours}} \\ \cline{2-9}
 & \textbf{TNG} & \textbf{SIMBA} & \textbf{TNG} & \textbf{SIMBA} & \textbf{TNG} & \textbf{SIMBA} & \textbf{TNG} & \textbf{SIMBA} \\ \hline
PSNR $\uparrow$ & 20.9 & 21.1 & 26.0 & 22.1 & 31.1 & 29.6 & \textbf{33.3} & \textbf{30.9} \\ \hline
SSIM $\uparrow$ & 0.443 & 0.437 & 0.703 & 0.565 & 0.894 & 0.855 & \textbf{0.927} & \textbf{0.884} \\ \hline
PCC $\uparrow$  & 0.189 & 0.197 & 0.832 & 0.616 & 0.924 & 0.863 & \textbf{0.975} & \textbf{0.930} \\ \hline
\textit{Held-out} $\balpha$ RMSE $\downarrow$  & 3.697 & 3.919 & 1.475 & 3.295 & \textbf{0.411} & 1.597 & 0.530 & \textbf{1.485} \\ \hline
\textit{Held-out} $\kappa$ RMSE $\downarrow$   & 0.0552 & 0.0521 & 0.0173 & 0.0169 & 0.0067 & 0.0073 & \textbf{0.0058} & \textbf{0.0067} \\ \hline
\textit{Held-out} $\bgamma$ RMSE $\downarrow$  & 0.0493 & 0.0437 & 0.0188 & 0.0217 & 0.0094 & 0.0174 & \textbf{0.0084} & \textbf{0.0151} \\ \hline
\end{tabular}
\vspace{-.15in}
\end{table}

\vspace{-.05in}
\paragraph{Quantitative results.} \tref{tab:quantitative-results} displays quantitative results comparing our method with Kaiser-Squires \cite{kaiser1993mapping}, \customcitet{Napier}{napier2023hubble}, and our UNet baseline. We report the median PSNR, Structure Similarity Index (SSIM), Pearson Correlation Coefficient (PCC), and the deflection angle $\balpha$, convergence $\kappa$, and shear $\bgamma$ RMSE at held-out points (i.e., points without photometry or lensing signals). We do so across 100 clusters by comparing each reconstructed mass map to its ground truth in \dataset. Our method outperforms all baselines on practically all metrics, both in our TNG test set (which uses the same simulation framework as the train set), and our SIMBA test set (which does not), demonstrating generalization towards an entirely unseen simulation environment.

\begin{figure}[t]
    \centering
    \captionsetup{skip=-4pt}
    \def\svgwidth{\textwidth} 
    \begin{small}
    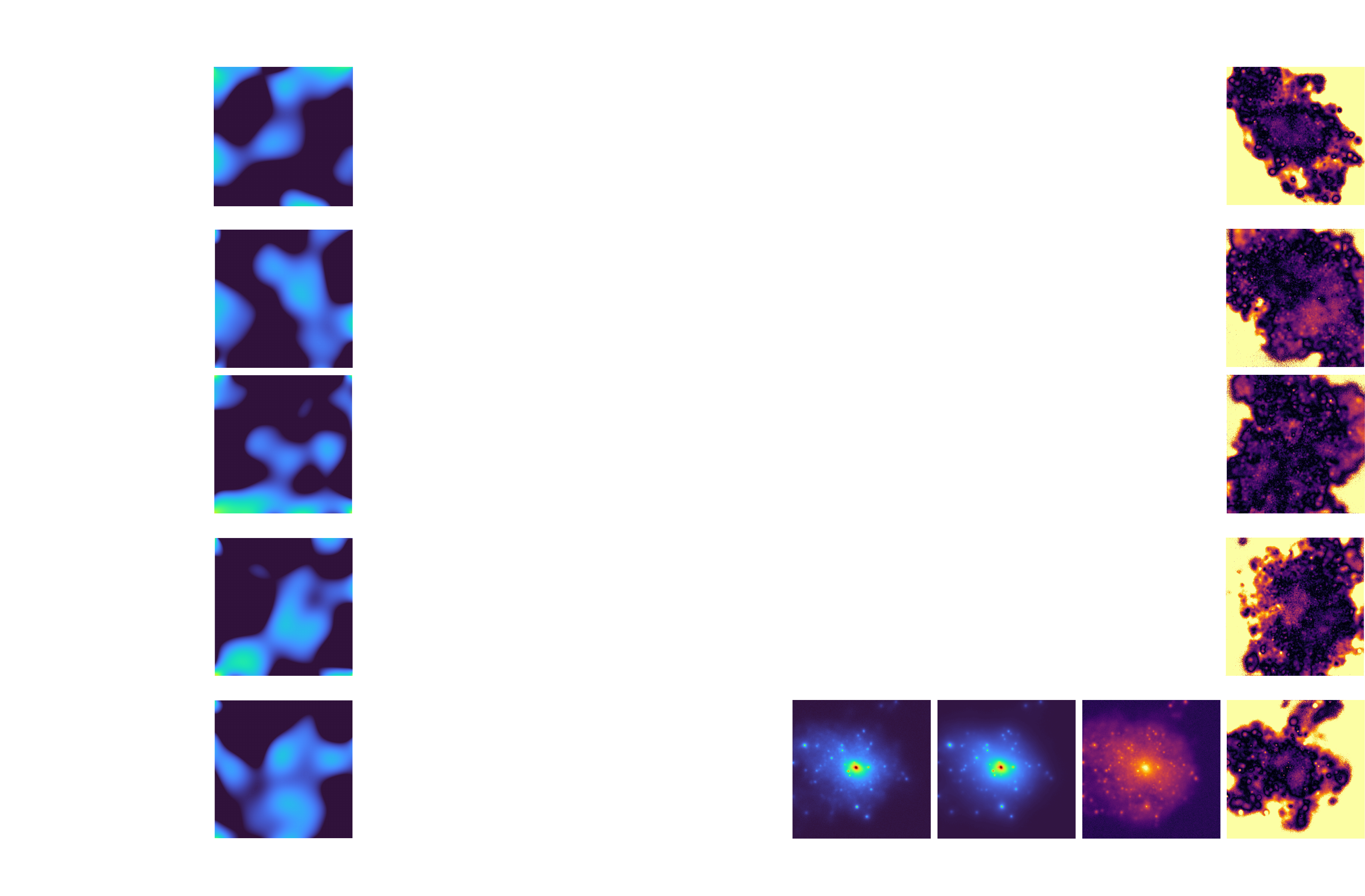
    \end{small}
    \caption{Comparison of our method with Kaiser-Squires (KS) \cite{kaiser1993mapping}, \customcitet{Napier}{napier2023hubble} and our UNet baselines. We show lens convergence $\kappa$ for five test samples of \dataset\ at different redshifts $z_L$ and simulations. Napier et al. produce an overly smooth mass prediction, \new{and yield inaccurate estimates in certain regions} due to its strict 1:1 mapping of mass and photometry. Also, contrary to both Napier et al. and our UNet baselines, our method includes gravitational lensing observables and generates samples from a posterior distribution with calibrated uncertainty estimation; we show one sample of this posterior along with the mean and std. dev. computed for 20 generated samples.
    }
    \label{fig:sextractor-comparison}
    \vspace{-.1in}
\end{figure}

\vspace{-.05in}
\paragraph{Qualitative results.} \fref{fig:sextractor-comparison} displays five galaxy clusters together with their reconstructed convergence maps $\kappa$. For each cluster, our method generates multiple reconstructions that integrate the learned prior with the lensing constraints. We report one representative sample, the posterior mean over 20 samples, and the standard deviation, which characterizes the uncertainty in the reconstruction.

We evaluate generalization at different redshifts and simulation frameworks. For this, we show reconstructions on our SIMBA test set, and for the IllustrisTNG test set we also show different distances $D_L$ of the lens plane, represented by their redshifts $z_L \in \{0.2, 0.5, 1.0\}$. Clusters at $z_L = 0.2$ are older and have had more time to develop complex structures, while clusters at $z_L = 1.0$ are younger and less evolved. Even though our method was trained only on clusters at $z_L = 0.5$ from IllustrisTNG, it generalizes well to clusters at different redshifts and simulated using different techniques.

The approach of \customcitet{Napier}{napier2023hubble} shows two main limitations. First, it overly smooths the mass distributions as SE++ uses one simple parametric model for each detected source, ignoring per-pixel differences in intensity. In contrast, our physically-guided diffusion pipeline captures these finer variations. Second, because the mass of each source is constrained to follow a 1:1 relation with light intensity, small deviations from this assumption can lead to significant overestimation (e.g., spurious features in the fourth and fifth rows) or underestimation (e.g., the outer regions of the first row) of the mass. Also, compared to our UNet baseline, our diffusion approach leverages gravitational lensing data to physically constrain the unseen dark matter, and produces uncertainty estimates to robustly quantify the model's confidence.

\vspace{-.05in}
\subsection{Mass reconstruction of MACS 1206}
\label{sec:real-cluster}

\new{As a proof of concept,} we further evaluate our physically guided diffusion pipeline outside of \dataset\ on the MACS 1206 galaxy cluster. This cluster, located at redshift $z_L \approx 0.44$, contains an unusually large number of multiply imaged background galaxies, which has allowed other works to study the mass distribution of this cluster in detail.
As detailed in the \supp, we use HST photometry $\Ib$ in the same three bands $\band \in \{125, 606, 814\}$ used to train our model, together with the strong lensing constraints from \customcitet{Caminha}{caminha2017mass}. %
\begin{figure}[t]
    \centering
    \captionsetup{skip=-6pt}
    \def\svgwidth{\columnwidth} 
    \begin{small}
    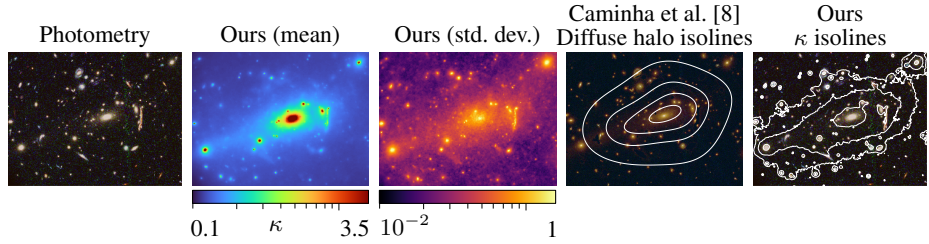
    \end{small}
    \caption{
    Our mass estimate for the MACS 1206 cluster, compared with the one shown in work by \customcitet{Caminha}{caminha2017mass}. We show our dimensionless surface mass density $\kappa$ estimate along with its isolines and critical curve at $z_S \approx 1.44$; the reference work shows the diffuse cluster halo which does not account for light sources in the lens plane and thus is naturally smoother.
    }
    \label{fig:real-cluster}
    \vspace{-.12in}
\end{figure}

In \fref{fig:real-cluster}, we compare our inferred surface mass density of MACS 1206 with the reconstruction using Lenstool from \customcitet{Caminha}{caminha2017mass}. We show the resulting dimensionless surface mass density $\kappa$ together with its isolines, and overlay the contours from Caminha et al. which trace the diffuse cluster halo, a smoother variant that does not account for light sources in the lens plane. Although the two reconstructions are derived under different assumptions, this side-by-side comparison highlights agreement in overall halo morphology and orientation, while small-scale peaks near individual cluster members naturally differ. \new{We also compare the critical curve at $z_S \approx 1.44$ (the curve marking the boundary of the strong-lensing region where multiple images appear), and find similar morphology. Our strong-lensing multiple-image Root Mean Square scatter, which measures the offset between the predicted and observed multiple-image positions, is 0.56 arcseconds. This is comparable to the 0.44 arcseconds of the expert-tuned reconstruction of \customcitet{Caminha}{caminha2017mass}. Finally, our estimate of the total projected mass enclosed within 100 kpc is $(7.61\pm 0.04) \times 10^{13}M_\odot$ across 20 generated samples. Compared to the estimate from Caminha et al., which is $(7.25\pm 0.02) \times 10^{13}M_\odot$, our enclosed mass error is just $5.03\% \pm 0.49\%$. As a reference, independent expert-tuned models of the same cluster routinely differ by $\sim\!10\%$ \cite{meneghetti2017frontier}.} Notably, our reconstruction requires no expert tuning, and matches expert-derived results in just a few minutes.%

\vspace{-.05in}
\section{Conclusion}
\label{sec:conclusion}
\vspace{-.03in}

We have introduced the first fully automatic framework for reconstructing galaxy cluster surface mass densities from both weak and strong lensing data. Central to this effort is \dataset, our new collection of \datasetsize\ \new{cluster observations} derived from IllustrisTNG and SIMBA, and rendered with realistic lensing observables, which enables training, benchmarking, and future community development. By coupling a diffusion prior that learns the joint statistics of mass and light with physically grounded forward models, our approach delivers accurate, high-resolution mass reconstructions in minutes without parametric assumptions or manual tuning. Together, our diffusion-based method and the \dataset\ benchmark establish a foundation for scalable, reproducible, and fully automated cluster mass inference, strengthening our tools for probing the Universe.

\section*{Acknowledgements}
This work has been supported by grant PID2022-141539NB-I00, funded by MICIU/AEI/10.13039/501100011033 and by ERDF, EU. It has also been supported by an Amazon AI4Science Discovery Award, and a Sloan Research Fellowship. Additionally, Diego Royo was supported by a Gobierno de Aragón predoctoral grant (CUS/803/2021) and mobility grant (ECU/1841/2023). We are grateful to Liam Connor and Samuel McCarty for their helpful discussions, to the anonymous reviewers for their time and insightful comments, and to Edurne Bernal-Berdún for her assistance with the figures.

\bibliographystyle{splncs04}
\bibliography{tex/11_references}

\newpage
\appendix
\section{Construction of \dataset}
\label{sec:appendix:clusters}

To construct our \dataset\ dataset, we mostly used the TNG300-1 subset of the IllustrisTNG cosmological simulation \cite{Pillepich_2017}. This dataset spans a cubic volume with a side length of 205,000 $\text{ckpc}/h$ that includes both baryonic matter (visible matter: stars, black holes and gas) and dark matter. TNG300-1 contains thousands of galaxy clusters resolved at high mass and spatial resolution. Additionally, we used the SIMBA simulations \cite{dave2019simba}, which are similar in terms of contents, volume, and resolution, but rely on different algorithms than IllustrisTNG. Our SIMBA test set allows us to evaluate generalization of our method under different simulation settings.

The surface mass density maps $\Sigma(\boldsymbol\theta)$ include both baryonic (visible) and dark matter, and are measured in $(10^{10}M_\odot/h)/(\text{ckpc}/h)^2$, i.e., in terms of $10^{10}$ times the mass of the Sun $M_\odot$ per comoving kiloparsec squared, all scaled by the Hubble parameter $h$. We also release maps where both components have been separated to facilitate analysis.  The photometry information $\Ib(\boldsymbol\theta)$ is measured in janskys (Jy). Since all \dataset\ images are $512\times512$ pixels, each pixel corresponds to an area of $1.56\;(\text{ckpc}/h)^2$ for images with a field of view of $100\times100$ arcseconds, and $7.87\;(\text{ckpc}/h)^2$ in the case of $225\times225$ arcseconds, for clusters at $z_L = 0.5$. \new{These correspond to image resolutions of 0.2 and 0.44 arcseconds respectively. For reference, dark matter distributions are substantially smoother, and existing mass reconstruction methods typically operate at resolutions of 0.5 to 1 arcsecond \cite{diego2025euclid, cha2022mars}.}

\begin{figure}[t]
    \centering
    \captionsetup{skip=-4pt}
    \def\svgwidth{\columnwidth} 
    \begin{small}
    \begingroup%
  \makeatletter%
  \providecommand\color[2][]{%
    \errmessage{(Inkscape) Color is used for the text in Inkscape, but the package 'color.sty' is not loaded}%
    \renewcommand\color[2][]{}%
  }%
  \providecommand\transparent[1]{%
    \errmessage{(Inkscape) Transparency is used (non-zero) for the text in Inkscape, but the package 'transparent.sty' is not loaded}%
    \renewcommand\transparent[1]{}%
  }%
  \providecommand\rotatebox[2]{#2}%
  \newcommand*\fsize{\dimexpr\f@size pt\relax}%
  \newcommand*\lineheight[1]{\fontsize{\fsize}{#1\fsize}\selectfont}%
  \ifx\svgwidth\undefined%
    \setlength{\unitlength}{448.78977378bp}%
    \ifx\svgscale\undefined%
      \relax%
    \else%
      \setlength{\unitlength}{\unitlength * \real{\svgscale}}%
    \fi%
  \else%
    \setlength{\unitlength}{\svgwidth}%
  \fi%
  \global\let\svgwidth\undefined%
  \global\let\svgscale\undefined%
  \makeatother%
  \begin{picture}(1,0.37515546)%
    \lineheight{1}%
    \setlength\tabcolsep{0pt}%
    \put(0,0){\includegraphics[width=\unitlength,page=1]{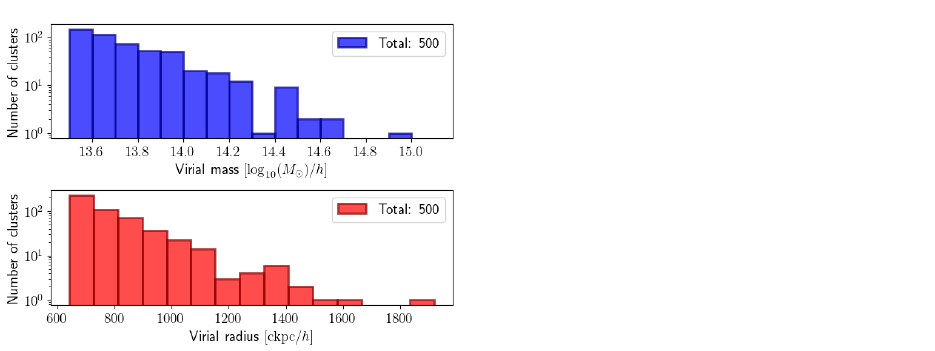}}%
    \put(0.24559128,0.36390997){\makebox(0,0)[t]{\lineheight{1.25}\smash{\begin{tabular}[t]{c}\textbf{IllustrisTNG}\end{tabular}}}}%
    \put(0.76230284,0.36390997){\makebox(0,0)[t]{\lineheight{1.25}\smash{\begin{tabular}[t]{c}\textbf{SIMBA}\end{tabular}}}}%
    \put(0,0){\includegraphics[width=\unitlength,page=2]{distribution-v2.pdf}}%
  \end{picture}%
\endgroup%

    \end{small}
    \caption{
    Distribution of the galaxy clusters selected from IllustrisTNG and SIMBA to construct \dataset. We show their virial masses (i.e., their total gravitationally-bound masses) and their virial radii (i.e., the radius that encloses that mass).
    }
    \label{fig:cluster_distribution}
    \vspace{-.1in}
\end{figure}

\paragraph{Selection of galaxy clusters.} TNG300-1 provides a friends-of-friends catalog of individual galaxy clusters within the simulation. For each identified cluster, we measure its total gravitationally-bound mass (the \emph{virial} mass) and the corresponding enclosing radius (the virial radius). We use a similar strategy for clusters from SIMBA, using their provided catalogs to filter by total cluster masses. To construct our dataset, we select 600 galaxy clusters with virial masses of at least $10^{13.5}$ times the mass of our Sun $M_\odot$. \fref{fig:cluster_distribution} shows the distribution of selected galaxy clusters by their virial masses and radii. As noted in our introduction, our analysis focuses on clusters in the range $10^{13.5\text{--}14} M_\odot$, which form the majority of our dataset.

\subsection{Parsing data from IllustrisTNG}
\label{sec:appendix:parsing}

Galaxy clusters contain volumetric information on mass and emitted light, represented as 3D particle clouds. Out of the 600 clusters in \dataset, 540 come from IllustrisTNG. In our work, we process this information to obtain 25 samples per cluster, where each sample consists of four images: one surface mass density map $\Sigma(\btheta)$, and three simulated HST images $\Ib$ (\sref{sec:dataset:generation}). Below, we describe the processing steps for each one.

\paragraph{Surface mass density maps $\Sigma(\btheta)$.} For baryonic matter, we use the \texttt{Coordinates} and \texttt{Masses} fields of gas, star and black hole particles. For dark matter, we use the \texttt{Coordinates} and \texttt{SubfindDensity} fields together with the \texttt{MassTable} attribute. This allows us to provide visible and dark matter components in \dataset, although in our work we use only their combined mass.

We generate 25 different 2D projections from each 3D particle cloud (\fref{fig:dataset_creation}). These correspond to 25 different viewing directions that are uniformly distributed over the surface of a hemisphere enclosing the cluster, each defined by a unit line-of-sight direction $\losdir_i$. In practice, each corresponds to a rotation of the particle cloud and a projection along $\losdir_i$. Then, to compute the mass at each pixel $\btheta$ in $\Sigma(\btheta)$, we find the 300 nearest particles and perform kernel density estimation with a box filter.

\paragraph{Photometry maps $\Ib$.} We use \texttt{Coordinates} and \texttt{GFM\_StellarPhotometrics} from star particles for each cluster. This last field contains stellar magnitudes in eight bands ($U$, $B$, $V$, $K$, $g$, $r$, $i$, $z$) \cite{Pillepich_2017}, which we convert to flux densities $f_\nu$ in Jy. As with the surface mass density maps, we generate the same projections by rotating the 3D particle cloud and projecting along each line-of-sight direction $\losdir_i$. Unlike the mass maps, we do not perform kernel density estimation, and instead we just bin each particle to its nearest pixel. To simulate HST observations, this binning accounts for dust attenuation and redshift, and then it uses the public response curves of the HST F125W, F606W and F814W wavelength filters for the bands $\band \in \{125, 606, 814\}$ respectively. This produces initial photometry maps, which we refine through post-processing as described below.

First, we add background light using publicly available mosaics from other HST captures.

Second, we add intracluster light (ICL), which corresponds to light from stars that are not bound to the galaxy cluster. Following \customcitet{Gonzalez}{gonzalez2005intracluster}, we model ICL using two Sérsic source parametric models \cite{sersic1963influence} fit to the brightest cluster galaxies. We use SE++ \cite{bertin1996sextractor} to do the fitting, then we create the Sérsic models and add the resulting light to the image.

Finally, we simulate several strongly lensed light sources. Following the process described in \sref{sec:dataset:generation} and \fref{fig:dataset_lensing}, we use a Sérsic parametric model for each light on the source plane $S$, and simulate the corresponding observed image on the lens plane $L$, which we add to the photometry.

\subsection{Parsing data from SIMBA}

Our pipeline for the SIMBA simulations \cite{dave2019simba} follows a very similar approach. We extract 60 galaxy clusters and also generate 25 projections per cluster, resulting in a test set of 1,500 samples. Just as with IllustrisTNG, each sample consists of one surface mass density map $\Sigma(\btheta)$ and three HST images $\Ib$ in identical units. The main differences from the IllustrisTNG setup are:

\paragraph{Surface mass density maps $\Sigma(\btheta)$.} While we also use the \texttt{Coordinates} and \texttt{Masses} for gas, stars and black holes, we now use the \texttt{Masses} field for dark matter particles as well.

\paragraph{Photometry maps $\Ib$.} Since SIMBA does not directly provide stellar magnitudes, we derive them from the \texttt{Masses}, \texttt{StellarFormationTime} and \texttt{Metallicity} fields of the star particles. We map these properties using stellar emission models to determine their base brightness, which we then scale by the particle's total mass to calculate the final magnitudes.

\section{Parsing data from MACS 1206}

To show that our method works on real observations, we applied it to the galaxy cluster MACS 1206. We chose this cluster since it is not affected by external convergence or shear (i.e., gravitational lensing deflection in the space before the cluster itself). Our estimation uses photometry $\Ib$ and strong lensing data, which we obtain as follows:

\paragraph{Photometry maps $\Ib$.} We use the publicly available CLASH drizzled
mosaics of MACS 1206 \cite{postman2012cluster}, taken in the same three filters as our synthetic data (F125W, F606W and F814W, i.e., $\band \in \{125, 606, 814\}$). We align the three mosaics and then crop and resample to $512\times512$ pixels, so that their field of view and resolution match those of \dataset. Finally, we convert the mosaics from their native units (electron counts per second) to flux densities $f_\nu$ in janskys (Jy), using the photometric keywords in the FITS headers.

\paragraph{Strong lensing constraints.} We extract strong lensing constraints from work by Caminha~et~al.~\cite{caminha2017mass}. From their catalog, we filter 55 multiple images that fall within the region covered by our photometry;
these belong to 15 distinct background sources, with redshifts ranging from
$z_S \approx 1.0$ to $z_S \approx 3.8$.

\section{Training and inference details}

For training and inference, we normalize the surface mass density maps $\Sigma$ to the $[0,1]$ range using a logarithmic transform followed by shift and scale operations. We apply the same normalization, independently, to the photometry maps $\Ib$.

Training follows default configuration provided by InverseBench \cite{zheng2025inversebench}. \new{Our 137M parameter model trains in around 10 hours on a single RTX 4090 GPU, where we run all of our experiments. All our experiments use a DDPM++ backbone. We also evaluated our method using a Diffusion Transformer backbone, which achieved comparable performance while requiring higher computational cost. This suggests we are not significantly bottlenecked by prior model choice in our experiments.}

During inference, the loss weights $\lambda_k$ for our strong and weak lensing forward operators in \eref{eq:losses} are set as follows. For strong lensing, the geometric and photometric losses use $\lambda_\text{s}^\text{geo} \approx 0.6 \cdot 10^{-2}$ and $\lambda_\text{s}^\text{img} \approx 0.6 \cdot 10^{-3}$ respectively, which puts both losses on roughly the same scale. For weak lensing, we use $\lambda_\text{w} \approx 0.05$. Additionally, for the DAPS algorithm, we set the parameters $\tau \approx 2.6 \cdot 10^{-3}$ and learning rate of $2.8\cdot10^{-6}$ over 100 steps. The rest of the parameters use the default values from InverseBench \cite{zheng2025inversebench}. All these values were chosen through parameter sweeps on a small validation subset of \dataset\ excluded from both training and final evaluation.

\section{Light-traces-mass in \dataset}

Here we measure the extent to which the light-traces-mass assumption holds in our \dataset\ dataset. There are several previous works \cite{napier2023hubble, kneib2011lenstool} that assume a 1:1 relation of mass to light, while our work incorporates a diffusion-based prior that can learn complex relations. For this, we compute the Pearson Correlation Coefficient (PCC) between the mass and photometry maps at the $\band = 125$ band, which we also use for our implementation of the method from \customcitet{Napier}{napier2023hubble} in \sref{sec:results:ourdataset}. We evaluate this metric on the three IllustrisTNG test subsets of \dataset, corresponding to galaxy cluster redshifts $z_L \in \{0.2, 0.5, 1.0\}$, obtaining values of 0.58, 0.61 and 0.65 respectively. These results indicate a moderate correlation. One of the main differences comes from the presence of massive clumps of dark matter which do not interact with light. There are also noticeable discrepancies primarily caused by background light, intracluster light (\sref{sec:appendix:parsing}) and the absence of the scaling relations from \customcitet{Limousin}{limousin2005constraining} (\sref{sec:results:ourdataset}).

Comparing these PCC values to those reported in \tref{tab:quantitative-results}, both the method of \customcitet{Napier}{napier2023hubble}, our UNet baseline, and, especially, our method achieve substantially higher correlations. As shown in \fref{fig:sextractor-comparison}, our approach learns how to effectively compensate for the mismatches mentioned before without assuming an explicit 1:1 mass-light relation, resulting in improved overall performance.

\end{document}